\documentclass{article}

\usepackage{microtype}
\usepackage{graphicx}
\usepackage{subcaption}
\usepackage{booktabs}
\usepackage{enumitem}

\usepackage{hyperref}

\usepackage[accepted]{icml2026_weightsymmetry}

\usepackage{amsmath}
\usepackage{amssymb}
\usepackage{mathtools}
\usepackage{amsthm}

\usepackage[capitalize,noabbrev]{cleveref}

\theoremstyle{plain}

\theoremstyle{definition}

\theoremstyle{remark}

\usepackage[textsize=tiny]{todonotes}

\icmltitlerunning{Analyzing Stream Collapse in Hyper-Connections: From Diagnosis to Mitigation}

\usepackage[dvipsnames]{xcolor}
\usepackage[most]{tcolorbox}
\usepackage{graphicx}
\usepackage{tabularx}
\usepackage{stfloats}
\usepackage{float}

\definecolor{HWPrimary}{RGB}{40, 90, 160}

\begin{document}

\twocolumn[
  \icmltitle{Analyzing Stream Collapse in Hyper-Connections: From Diagnosis to Mitigation
}

  \icmlsetsymbol{equal}{*}

  \begin{icmlauthorlist}
    \icmlauthor{Ekaterina Alimaskina}{equal,mirai,brain,yr}
    \icmlauthor{Gleb Molodtsov}{equal,mirai,brain}
    \icmlauthor{Aleksandr Beznosikov}{mirai,brain,ui}
  \end{icmlauthorlist}

  \icmlaffiliation{mirai}{MIRAI, Moscow, Russia}
  \icmlaffiliation{brain}{BRAIn Lab, Moscow, Russia}
  \icmlaffiliation{ui}{Innopolis University, Innopolis, Russia}
  \icmlaffiliation{yr}{Yandex Research, Moscow, Russia}

  \icmlcorrespondingauthor{Ekaterina Alimaskina}{alimaskina.ea@yandex.ru}

  \icmlkeywords{Weight-Space Symmetries, Quantization, Model Compression}

  \vskip 0.3in
]

\printAffiliationsAndNotice{}
\begin{abstract}
Hyper-Connections (HC) replace the single Transformer residual stream with multiple streams, introducing a permutation symmetry over stream indices.
We study how this symmetry is resolved in practice: whether streams specialize in a balanced way or exhibit dominant-stream usage.
Using fine-grained diagnostics for HC-based language models, we trace how multi-stream representations are actually used. We find that after an early seeding stage, residual mixing often remains close to identity, limiting a core HC mechanism for exchanging information between streams. 
Moreover, both signal and interpretable features concentrate in a dominant stream, and the nominally multi-stream residual connection can underutilize its capacity, behaving closer to a single-stream residual pathway.
Finally, we show that breaking symmetry at stream initialization reduces dominant behavior and improves performance across \textit{m}HC variants. Our code is publicly available\begingroup
\renewcommand{\thefootnote}{\fnsymbol{footnote}}\footnote[2]{\url{https://github.com/brain-lab-research/hc-stream-collapse}}\endgroup.
\end{abstract}

\section{Introduction}
\label{sec:intro}

Weight-space symmetries arise when different parameter configurations represent the same network function. While often viewed as harmless redundancies, they can also shape optimization by making some solutions easier to reach than others.
Prior work has often focused on symmetries inside Transformer blocks, such as those involving attention heads or feed-forward layers. In contrast, the symmetries introduced by residual pathways remain underexplored.

Each Transformer block applies attention and feed-forward transforms to a shared residual stream, producing an update:
\begin{equation}
\mathbf{x}_{\ell+1}
=
\mathbf{x}_{\ell}
+
\mathcal{F}_{\ell}(\mathbf{x}_{\ell};\mathbf{W}_{\ell}),
\label{eq:residual}
\end{equation}
where $\mathbf{x}_{\ell}$ denotes the residual representation at layer $\ell$, and $\mathcal{F}_{\ell}$ denotes the layer transformation.

Hyper-Connections (HC) replace one residual stream with $n$ parallel ones and learn token-dependent connectivity between them \citep{hc}. For one token, let $\mathbf{X}_{\ell}\in\mathbb{R}^{n\times d}$ denote the stream state at layer $\ell$, with rows corresponding to streams. An HC layer is
\begin{equation}
\mathbf{X}_{\ell+1}
=
\mathbf{H}^{\mathrm{res}}_{\ell}\mathbf{X}_{\ell}
+
\big(\mathbf{H}^{\mathrm{post}}_{\ell}\big)^{\!\top}
\mathcal{F}_{\ell}\!\left(
\mathbf{H}^{\mathrm{pre}}_{\ell}\mathbf{X}_{\ell};
\mathbf{W}_{\ell}
\right).
\label{eq:hc_layer}
\end{equation}

Here the connectivity operators are token-dependent: $\mathbf{H}^{\mathrm{res}}_{\ell}\in\mathbb{R}^{n\times n}$ mixes the carried stream state in the residual branch, while $\mathbf{H}^{\mathrm{pre}}_{\ell},\mathbf{H}^{\mathrm{post}}_{\ell}\in\mathbb{R}^{1\times n}$ form the read/write interface. The former reads from streams into the block, and the latter writes the block update back. When $n=1$, Eq.~\eqref{eq:hc_layer} reduces to Eq.~\eqref{eq:residual}.

By construction, HC first expands the residual representation into $n$ streams. In the standard setup, this is done by replication, $\mathbf{x}\mapsto[\mathbf{x};\ldots;\mathbf{x}]$, so all streams start as identical copies. Together with Eq.~\eqref{eq:hc_layer}, this creates a permutation symmetry over stream indices: relabeling the streams and corresponding connectivity operators preserves the represented computation. This raises a practical question:
\begin{center}
\textit{If streams start identical and exchangeable, what makes them acquire different roles during training, and do they actually specialize in practice?}
\end{center}

Prior HC variants primarily focus on architecture design, stability, and aggregate language-modeling performance, leaving the internal use of individual streams less characterized. In particular, they address the instability of unconstrained residual mixing: unrolling Eq.~\eqref{eq:hc_layer} replaces the fixed identity path by products of $\mathbf{H}^{\mathrm{res}}_{\ell}$ matrices, which may cause residual norms to grow or decay rapidly. Manifold-Constrained Hyper-Connections (\textit{m}HC) therefore constrain $\mathbf{H}^{\mathrm{res}}_{\ell}$ to be approximately doubly stochastic \citep{mhc}, i.e., to lie in the Birkhoff polytope
\begin{equation}
\mathcal{B}_{n}
=
\left\{
\mathbf{H}\in\mathbb{R}^{n\times n}
\,\middle|\,
\mathbf{H}\mathbf{1}=\mathbf{1},
\mathbf{1}^{\top}\mathbf{H}=\mathbf{1}^{\top},
\mathbf{H}\ge 0
\right\}.
\label{eq:birkhoff}
\end{equation}
This can be interpreted as mixing streams by convex combinations of permutations \citep{birkhoff1946tres}. Efficient variants such as \textit{m}HC-lite and KromHC provide cheaper constrained parameterizations \citep{mhc-lite,kromhc}.

These constraints address the stability of residual mixing, but they do not show whether the learned multi-stream mechanism is used as intended. In particular, the read/write interface remains a separate degree of freedom, and the standard symmetric expansion provides no controlled way to assign stream roles.
This creates a possible failure mode: a small early advantage for one stream may be amplified across depth, as that stream grows in norm, contributes more to later reads, and receives more subsequent updates. Rather than producing balanced specialization, training may therefore favor a dominant-stream regime.

We find this collapse pattern across our main HC experiments.
Read/write signal, representation norms, and semantic probes identify one dominant stream, while residual mixing often stays close to identity. To test whether this failure mode can be mitigated by controlled symmetry breaking, we study Learned Stream Scaling (LSS), a minimal modification at the expansion interface. LSS replaces exact stream replication with learnable near-identity diagonal scales, adding only $nd$ parameters while leaving the core HC operator unchanged. This helps the model to drift away from the dominant-stream regime and improves perplexity.

\vspace{-3mm}
\paragraph{Contributions.}
Our contributions are:
\begin{itemize}[leftmargin=*,nosep]
  \item We identify a stream-level failure mode in HC-style residuals: models with multiple symmetric streams can rely on one dominant stream.
  \item We show that collapse arises in mechanics and semantics: residual mixing stays near identity, while read/write signal and representation content concentrate in one stream.
  \item Using Learned Stream Scaling, we show that a small controlled symmetry break reduces collapse and improves \textit{m}HC variants without changing the core HC operator.
\end{itemize}

\section{Stream collapse under symmetric HC initialization}
\label{sec:collapse}

\subsection{Experimental protocol}
\label{sec:protocol}

We analyze \textit{m}HC-lite nanoGPT models \citep{nanogpt} with $n=4$ streams, following prior HC work \citep{mhc, mhc-lite}. See Appendix~\ref{sec:app_stream_counts} for other stream counts. Main figures use the medium model trained on OpenWebText. We evaluate perplexity on OpenWebText, WikiText-103, and C4 \citep{openwebtext,wikitext,c4}. See details in Appendix~\ref{sec:app_details}.

\subsection{Residual mixing is mostly near-identity}
\label{sec:width_mixing}

Prior HC variants often attribute their gains to the residual mixing operator $\mathbf{H}^{\mathrm{res}}_{\ell}$: ablations indicate that the largest improvements come from allowing width-wise exchange between streams \citep{mhc}. If residual mixing were the main driver of HC gains in the trained model, we would expect substantial cross-stream exchange beyond the earliest layers. We find that this is mostly not the case.

\begin{figure}[htbp]
  \centering
  \vspace{-4mm}
  \includegraphics[width=\columnwidth]{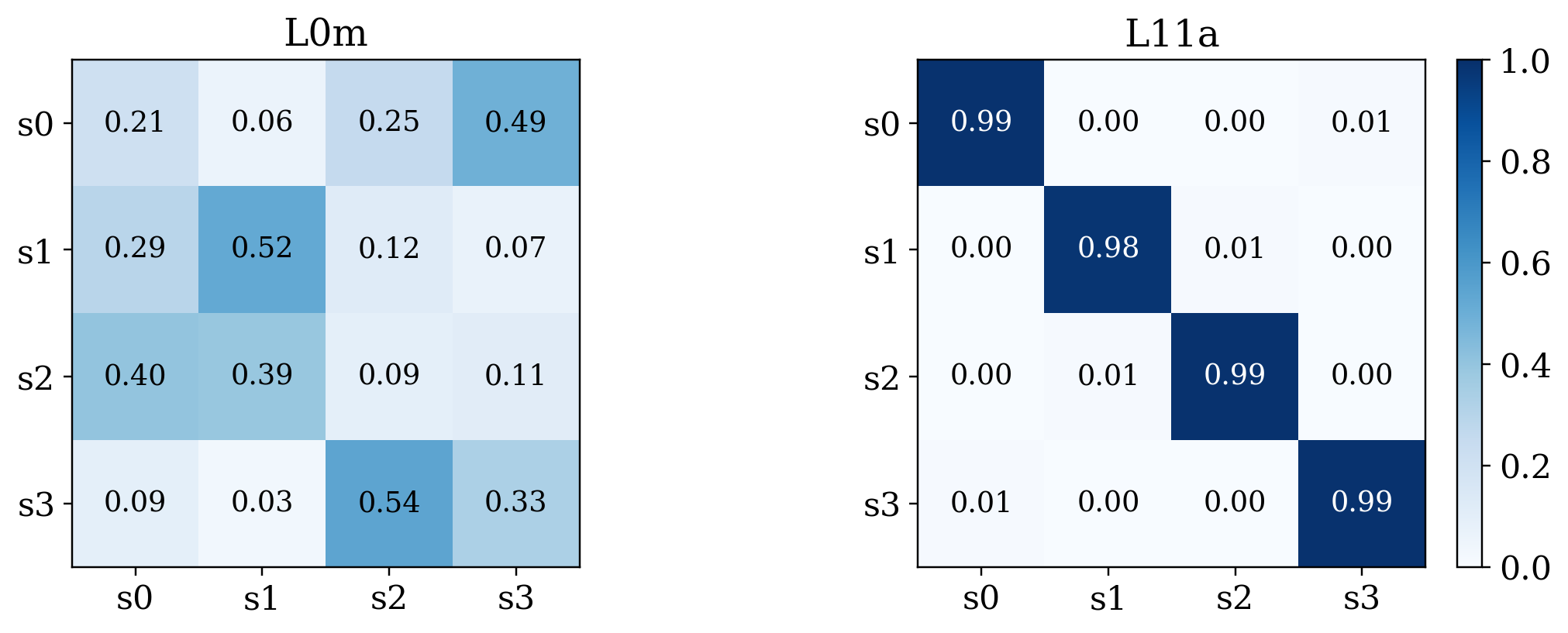}
  \caption{Token-averaged residual-mixing matrices $\mathbf{H}^{\mathrm{res}}_{\ell}$ in trained
\textit{m}HC-lite. Labels $s0$--$s3$ denote streams. The early mixer (\texttt{L0m}, left) has substantial off-diagonal mass, while a deeper mixer (\texttt{L11a}, right) is close to identity.} 
\label{fig:hres_mixing_examples}
  \vspace{-2mm}
\end{figure}

As shown in Fig.~\ref{fig:hres_mixing_examples}, the early mixer has substantial off-diagonal mass, while a deeper mixer is close to identity, suggesting that residual mixing becomes a near-bypass after initial seeding; full depth-wise visualizations are provided in Appendix~\ref{sec:app_mixing}.

A possible interpretation is that early updates differentiate the initially identical streams, so residual mixing is most useful at this stage. Once read/write usage concentrates on one stream, later cross-stream mixing may become less beneficial, consistent with the observed shift from early mixing to near-identity $\mathbf{H}^{\mathrm{res}}_{\ell}$ matrices.

To test whether sustained residual mixing is critical under the default regime, we freeze most $\mathbf{H}^{\mathrm{res}}_{\ell}$ matrices to identity and leave only every sixth layer trainable.\footnote{The strongest mixing appears early, but we keep every sixth mixer trainable to allow limited later exchange.}
If cross-stream exchange were a key performance driver, this intervention should substantially degrade perplexity. Instead, performance is mostly preserved, and even improves for the medium model (Table~\ref{tab:identity_hres}). This suggests that baseline models often operate near a bypass regime, where much of the residual mixing mechanism is underutilized.

\begin{table}[htbp]
\centering
\vspace{-2mm}
\captionsetup{width=\columnwidth,font=small}
\caption{\textit{m}HC-lite vs \textsc{Identity} residual mapping.
Perplexity ($\downarrow$) on validation/test sets.}
\vspace{-2mm}
\label{tab:identity_hres}
\small
\setlength{\tabcolsep}{4pt}
\begin{tabular}{llrrrr}
\toprule
Size & Model & OWT val & WT val & WT test & C4 test \\
\midrule
M & \textit{m}HC-lite & 26.22 & 52.74 & 51.05 & 57.32 \\
M & Identity $\mathbf{H}^{\mathrm{res}}_{\ell}$ & \textbf{25.93} & \textbf{51.86} & \textbf{49.65} & \textbf{53.94} \\
\midrule
L & \textit{m}HC-lite & \textbf{24.51} & \textbf{46.03} & \textbf{43.98} & \textbf{50.95} \\
L & Identity $\mathbf{H}^{\mathrm{res}}_{\ell}$ & 24.54 & 46.30 & 44.32 & 55.15 \\
\bottomrule
\end{tabular}
\end{table}
\vspace{-1mm}
\begin{tcolorbox}[
  colback=white,
  colframe=Black!50!white,
  boxrule=0.8pt,
  arc=2mm,
  left=2mm,right=2mm,top=1.5mm,bottom=1.5mm
]
\textbf{Takeaway.}
After an early seeding stage, residual mixing often stays close to identity. Stream indices therefore remain approximately aligned across layers, making it meaningful to track per-stream read/write usage, norm growth, and semantic content below.
\end{tcolorbox}

\subsection{Read/write routing favors one stream}
\label{sec:readwrite}

Beyond residual mixing, hyper-connections introduce a read-compute-write interface. If multiple streams are used as intended, then
several streams should contribute to the block input and receive meaningful updates across depth.

\begin{figure}[htbp]
  \centering
  \includegraphics[width=\columnwidth]{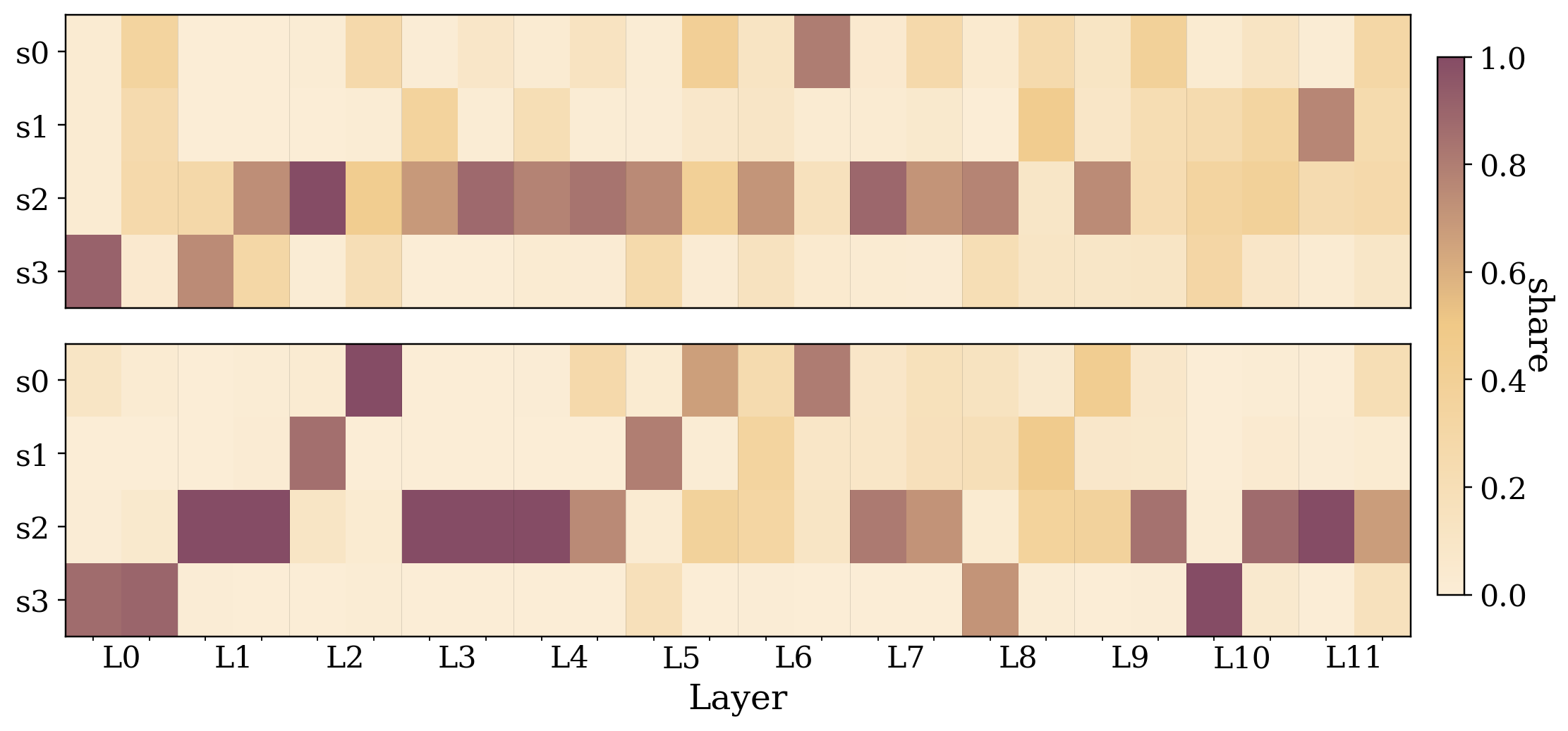}
  \vspace{-4mm}
  \caption{Token-averaged, normalized stream contributions in the read/write interface ($n=4$ streams). \textbf{Top:} average contribution of each stream to the block input (read). \textbf{Bottom:} average fraction of the block update
  written back to each stream (write). Stream \(s_2\) dominates both.}
\label{fig:read_write_share}
  \vspace{-2mm}
\end{figure}

Figure~\ref{fig:read_write_share} reports per-layer average stream
contributions to the read vector $\mathbf{H}^{\mathrm{pre}}_{\ell}$ and the write vector $\mathbf{H}^{\mathrm{post}}_{\ell}$. A single stream quickly becomes dominant: it provides most of the block input and receives most of the block update. Thus, the block is repeatedly read from and written back to one persistent stream, while the remaining streams receive substantially weaker update signal.

Thus, collapse is visible not only in residual mixing but also at the block interface: input and update signals are repeatedly routed through the same stream. We next show that this read/write imbalance is accompanied by a depth-wise norm imbalance.

\subsection{Depth-wise accumulation amplifies the dominant stream}
\label{sec:depth_collapse}

The read/write imbalance above suggests a simple mechanism for signal accumulation. In HC-style blocks, updates are added to streams across depth. If the write operator repeatedly favors one stream, that stream receives larger effective increments and its norm grows faster. A larger-norm stream can then contribute more to later block inputs through both learned read weights and activation scale, reinforcing its advantage over depth. This creates a self-reinforcing loop that amplifies an initially uneven read/write pattern into a dominant-stream regime.

\begin{figure}[htbp]
  \centering
  \includegraphics[width=\columnwidth]{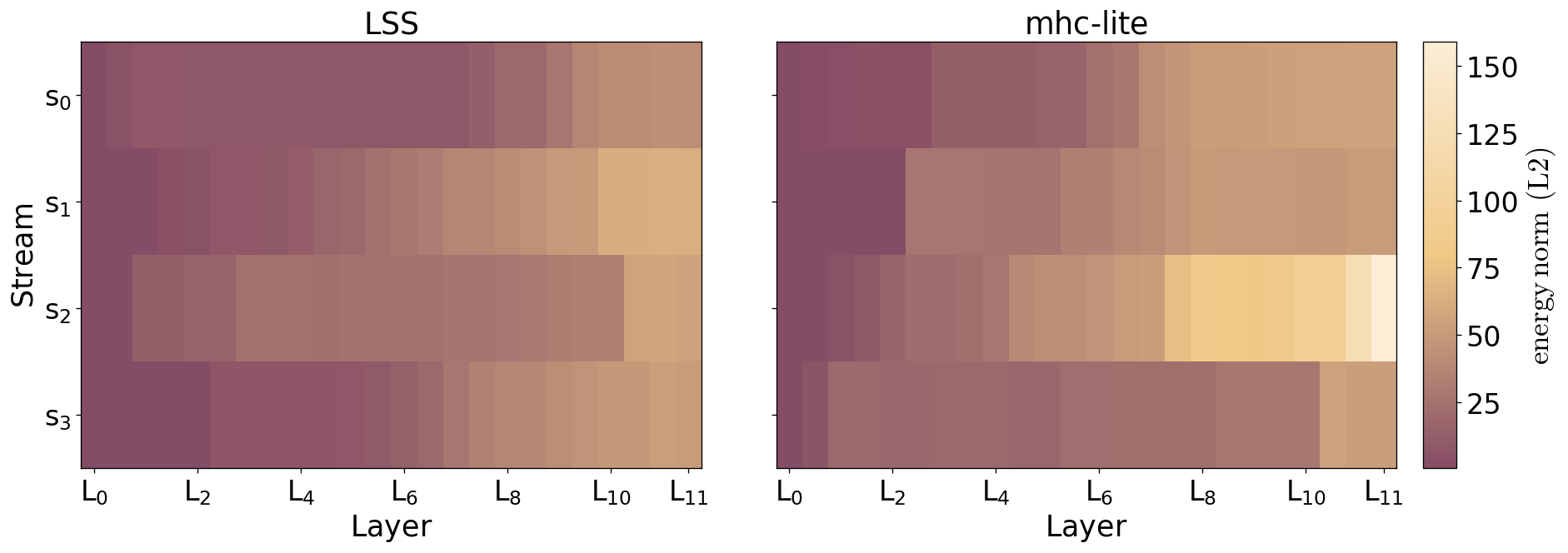}
  \vspace{-4mm}
  \caption{Per-stream representation $L_2$ norm, averaged across validation tokens. The \textit{m}HC-lite baseline shows a strong norm imbalance, while the LSS variant, introduced in Sec.~\ref{sec:lss}, reduces this imbalance.}
  \label{fig:norms_lss}
  \vspace{-6mm}
\end{figure}

Figure~\ref{fig:norms_lss} shows the corresponding representation-level imbalance for the \textit{m}HC-lite baseline. In the right panel, one stream accumulates markedly higher representational \(L_2\) norm in deeper layers, while the other streams carry much lower \(L_2\) norm. This provides a representation-level signature of the same collapse as at the read/write interface.

\subsection{Dominance extends to semantic representation structure}
\label{sec:semantic}

We now ask whether the same stream also carries more stable and interpretable structure. We use two complementary semantic probes: residual curvature and sparse crosscoders. These tools test token-level stream dominance via smoother residual trajectories and through crosscoder features that are preferentially expressed in a particular stream.

\begin{figure}[htbp]
  \centering
  \includegraphics[width=0.98\columnwidth]{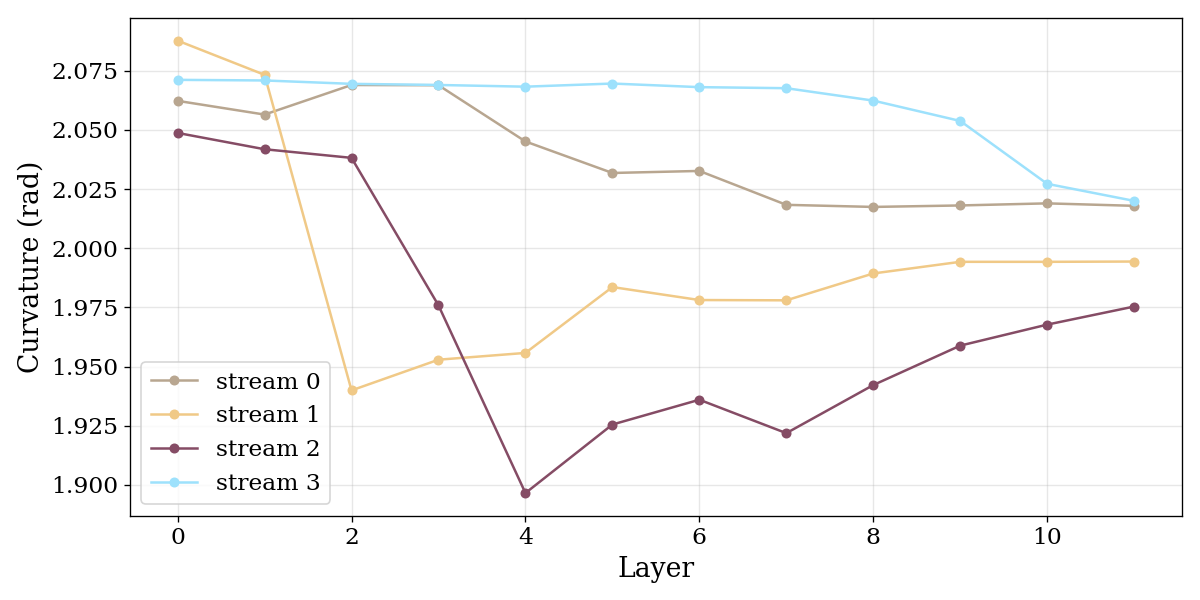}
  \caption{Token-trajectory curvature per stream across layers. Lower curvature indicates smoother token-level representation geometry. The stream that dominates the read/write interface also exhibits the lowest curvature over many layers.}
  \label{fig:curvature}
  \vspace{-4mm}
\end{figure}

\paragraph{Probe 1: residual curvature.}
Residual curvature measures how smoothly representations evolve along the token sequence. Following the curved-inference view of language models \citep{curvative}, we compute the average turning angle between consecutive token-to-token displacement vectors in hidden-state space. Lower curvature indicates a smoother and more stable representation trajectory; prior work links such measures to representation quality and downstream behavior \citep{skean2025layer}. We use this probe to test whether the read/write-dominant stream is also a geometrically more stable stream. Figure~\ref{fig:curvature} shows the same trend: the dominant stream has lower curvature over many layers, suggesting that it is not only larger in norm or preferred by the interface, but also carries a more stable representational signal.

\begin{table*}[t]
\centering
\caption{Comparison of \textit{m}HC and \textit{m}HC-lite variants.
Perplexity ($\downarrow$). \textsc{Identity} replaces all residual mixing
matrices by identity except every sixth layer.}
\label{tab:mhc_mhclite_variants}
\scriptsize
\setlength{\tabcolsep}{3pt}
\renewcommand{\arraystretch}{1.05}
\resizebox{\textwidth}{!}{
\begin{tabular}{l|cc|cc|cc|cc|cc|cc|cc|cc}
\toprule
& \multicolumn{8}{c|}{\textbf{\textit{m}HC}}
& \multicolumn{8}{c}{\textbf{\textit{m}HC-lite}} \\
\cmidrule(lr){2-9}\cmidrule(lr){10-17}
& \multicolumn{2}{c|}{OWT val}
& \multicolumn{2}{c|}{WT val}
& \multicolumn{2}{c|}{WT test}
& \multicolumn{2}{c|}{C4 test}
& \multicolumn{2}{c|}{OWT val}
& \multicolumn{2}{c|}{WT val}
& \multicolumn{2}{c|}{WT test}
& \multicolumn{2}{c}{C4 test} \\
\cmidrule(lr){2-3}\cmidrule(lr){4-5}
\cmidrule(lr){6-7}\cmidrule(lr){8-9}
\cmidrule(lr){10-11}\cmidrule(lr){12-13}
\cmidrule(lr){14-15}\cmidrule(lr){16-17}
Method
& M & L & M & L & M & L & M & L
& M & L & M & L & M & L & M & L \\
\midrule
Baseline
& 26.44 & 24.61 & 53.29 & 46.35 & 50.97 & 44.38 & 58.62 & 55.39
& 26.22 & 24.51 & 52.74 & 46.03 & 51.05 & 43.98 & 57.32 & \textbf{50.95} \\
Identity
& 26.72 & 24.45 & 53.15 & 46.07 & 50.88 & 43.95 & 57.72 & 51.38
& 25.93 & 24.54 & 51.86 & 46.30 & 49.65 & 44.32 & \textbf{53.94} & 55.15 \\
LSS $+$ Identity
& \textbf{25.74} & 24.35 & 51.54 & 45.87 & \textbf{48.92} & 43.80 & 55.16 & 53.37
& 25.88 & 24.36 & 51.96 & 44.70 & 49.60 & 42.69 & 57.04 & 52.18 \\
LSS
& 26.08 & \textbf{22.78} & \textbf{51.43} & \textbf{45.73} & 48.99 & \textbf{43.66} & \textbf{54.91} & \textbf{49.14}
& \textbf{25.65} & \textbf{24.29} & \textbf{49.41} & \textbf{44.64} & \textbf{47.92} & \textbf{42.62} & 54.20 & 51.23 \\
\bottomrule
\end{tabular}
}
\vspace{-5mm}
\end{table*}

\paragraph{Probe 2: sparse crosscoders.}
As a second probe, we train sparse crosscoders on hidden states from all four streams at even layers \(\{0,2,4,6,8,10\}\) \citep{crosscoder}. Sparse dictionary methods decompose representations into sparse, often interpretable feature directions; crosscoders learn one shared sparse code for all stream slices at a layer, making them suitable for localizing such features across streams. We assign each recovered feature to the stream with the largest decoder norm; details are given in Appendix~\ref{sec:app_crosscoder}.

Figure~\ref{fig:crosscoder_even} shows that recovered feature mass is highly concentrated. Early layers are dominated by \(s_3\), but from layer 4 onward most feature mass concentrates in \(s_2\), matching the stream identified by the read/write and norm probes. This provides feature-level evidence that the dominant stream also captures much of the recovered interpretable structure.

\begin{tcolorbox}[
  colback=white,
  colframe=Black!50!white,
  boxrule=0.8pt,
  arc=2mm,
  left=2mm,right=2mm,top=1.5mm,bottom=1.5mm
]
\textbf{Takeaway.}
Mechanical and semantic probes consistently identify the same dominant pathway, suggesting that residual capacity is not evenly distributed across streams.
\end{tcolorbox}

\section{Controlled initial symmetry breaking with Learned Stream Scaling}
\label{sec:lss}

The collapse above suggests a simple test: if exact stream replication contributes to uncontrolled role assignment, then a small perturbation of the replicated initialization may move the model away from the dominant-stream regime. Learned Stream Scaling (LSS) applies this break only at the expansion interface, leaving the HC operator unchanged.

\subsection{LSS breaks initial exchangeability}
\label{sec:lss_breaks}

The HC update is equivariant to stream relabeling. For a fixed token, suppressing token dependence, let $\mathbf{P}$ be a permutation matrix acting on the stream axis and define
\begin{equation}
\begin{aligned}
\mathbf{X}'_{\ell} &= \mathbf{P}\mathbf{X}_{\ell},
&
\mathbf{H}_{\ell}^{\prime\,\mathrm{res}}
&= \mathbf{P}\mathbf{H}^{\mathrm{res}}_{\ell}\mathbf{P}^{\top},
\\
\mathbf{H}_{\ell}^{\prime\,\mathrm{pre}}
&= \mathbf{H}^{\mathrm{pre}}_{\ell}\mathbf{P}^{\top},
&
\mathbf{H}_{\ell}^{\prime\,\mathrm{post}}
&= \mathbf{H}^{\mathrm{post}}_{\ell}\mathbf{P}^{\top}.
\end{aligned}
\label{eq:stream_relabeling}
\end{equation}
Substituting these into Eq.~\eqref{eq:hc_layer} gives $\mathbf{X}'_{\ell+1}=\mathbf{P}\mathbf{X}_{\ell+1}$, so stream labels have no intrinsic roles: relabeling streams together with their connectivity operators preserves the represented computation.

This symmetry becomes a practical issue because the standard HC expansion also gives streams no initial differences: it simply replicates the same representation, $\mathbf{E}(\mathbf{x})=[\mathbf{x};\ldots;\mathbf{x}]$. Since $\mathbf{P}\mathbf{E}(\mathbf{x})=\mathbf{E}(\mathbf{x})$ for any permutation $\mathbf{P}$, all streams start in the fixed subspace of the symmetry. Thus, role separation must emerge from training dynamics rather than from a controlled signal.

LSS targets this point by replacing exact replication with near-identity per-stream diagonal scaling:
\begin{equation}
\mathbf{x}^{(s)}
=
\mathbf{x}\odot\mathbf{h}_{s},
\qquad
\mathbf{h}_{s}\in\mathbb{R}^{d},
\qquad
s=1,\ldots,n.
\label{eq:lss}
\end{equation}
The scales are initialized near uniform, so streams remain close but are no longer identical. This gives streams distinct initial signals while leaving the HC update unchanged, adding only $nd$ parameters and no extra matmuls.
\subsection{LSS changes the stream-usage regime}
\label{sec:lss_mixing}

We test whether LSS moves the model away from the near-bypass regime.
First, we repeat the identity-mixing ablation from Sec. \ref{sec:width_mixing}, fixing all residual mixing matrices to identity except every sixth layer.
The baseline is almost insensitive to this ablation, consistent with near-bypass behavior. With LSS, the same constraint tends to be more costly, especially in larger models, suggesting that cross-stream mixing becomes more relevant under the LSS regime.

Second, we directly examine the collapse signature.
Figure \ref{fig:norms_lss} shows that LSS substantially reduces the strong norm imbalance observed in the baseline: the representation norm is less concentrated in a single stream.
Thus, LSS appears to change the stream-usage regime, rather than only improving perplexity within the same collapsed solution.
Additional scaling results across token budgets are provided in App.~\ref{sec:app_scaling}.

\section{Conclusion}
We show that HC can fail to realize their intended multi-stream behavior. After early mixing, trained models concentrate most signal, representation norm, and interpretable features in a single dominant stream. This suggests that stream-permutation symmetry may bias optimization toward collapsed residual pathways. A minimal symmetry break, Learned Stream Scaling, reduces this collapse and improves perplexity across \textit{m}HC variants. 





\bibliography{references}
\bibliographystyle{icml2026}


\newpage
\appendix
\onecolumn
\allowdisplaybreaks

\section*{Supplementary Material}

\vspace{4mm}

\section{Sparse crosscoder details}
\label{sec:app_crosscoder}

Sparse crosscoders are designed to compare representations across multiple sources using a shared sparse dictionary \citep{crosscoder}. This makes them a natural tool for our setting: instead of training separate sparse autoencoders for each stream, we train a crosscoder that reconstructs all stream slices at a given layer from the same sparse activation vector. Thus, a recovered feature can be compared across streams through its stream-specific decoder vectors.

For each selected even layer \(\ell\in\{0,2,4,6,8,10\}\), we collect hidden states
\[
\mathbf{x}^{(\ell,s)}_t \in \mathbb{R}^{d},
\qquad
s\in\{0,1,2,3\},
\]
where \(t\) indexes tokens and \(s\) indexes streams. The crosscoder learns non-negative sparse activations \(\mathbf{a}^{(\ell)}_t\in\mathbb{R}^{m}_{\ge 0}\) and stream-specific decoders \(\mathbf{D}^{(\ell,s)}\in\mathbb{R}^{d\times m}\), reconstructing each stream slice as
\[
\widehat{\mathbf{x}}^{(\ell,s)}_t
=
\mathbf{D}^{(\ell,s)}\mathbf{a}^{(\ell)}_t.
\]
The same sparse code \(\mathbf{a}^{(\ell)}_t\) must reconstruct all streams at that layer, so the learned features are aligned across streams rather than fitted independently.

To localize feature \(i\), we use the norm of its decoder column in each stream:
\[
w_i^{(\ell,s)}
=
\left\|
\mathbf{D}^{(\ell,s)}_{:,i}
\right\|_2 .
\]
Following standard crosscoder attribution, feature \(i\) is assigned to the stream with the largest decoder norm,
\[
s_i^{(\ell)}
=
\arg\max_s w_i^{(\ell,s)}.
\]
We then aggregate these assignments to obtain the feature share of each stream at layer \(\ell\).

\begin{figure}[H]
  \centering
  \includegraphics[width=\linewidth]{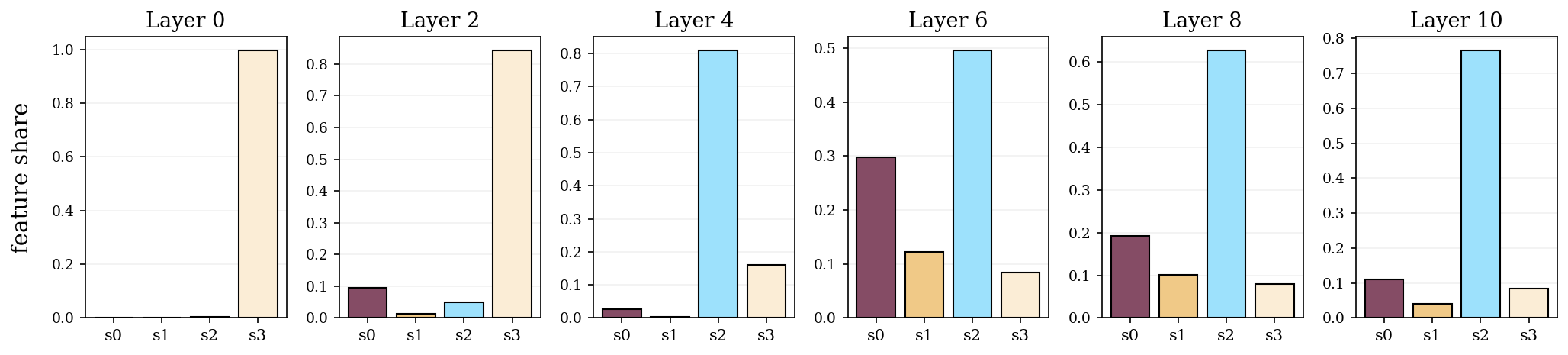}
  \caption{Sparse crosscoder trained on stream states from even layers.
  Recovered sparse-feature assignments concentrate in one stream at a time, providing interpretable evidence of dominant-stream behavior under HC-style residuals.}
  \label{fig:crosscoder_even}
\end{figure}

In Fig.~\ref{fig:crosscoder_even}, we report the stream-level feature share across selected even layers. The earliest layers are dominated by \(s_3\), but from layer 4 onward the recovered features concentrate in \(s_2\). This agrees with the read/write and norm probes in the main text, indicating that the dominant stream also carries most localized sparse features after the initial transition.

\newpage

\section{Stream count analysis}
\label{sec:app_stream_counts}

Prior HC work commonly uses \(n=4\) streams, but stream collapse may depend on
the number of available streams. We therefore repeat our diagnostics for
\(n\in\{2,4,8,16\}\). For this sweep, we use \textit{m}HC rather than
\textit{m}HC-lite, since the \textit{m}HC-lite permutation-based
parameterization scales factorially with the number of streams.

\begin{figure*}[!htbp]
  \centering
  \includegraphics[width=0.78\textwidth]{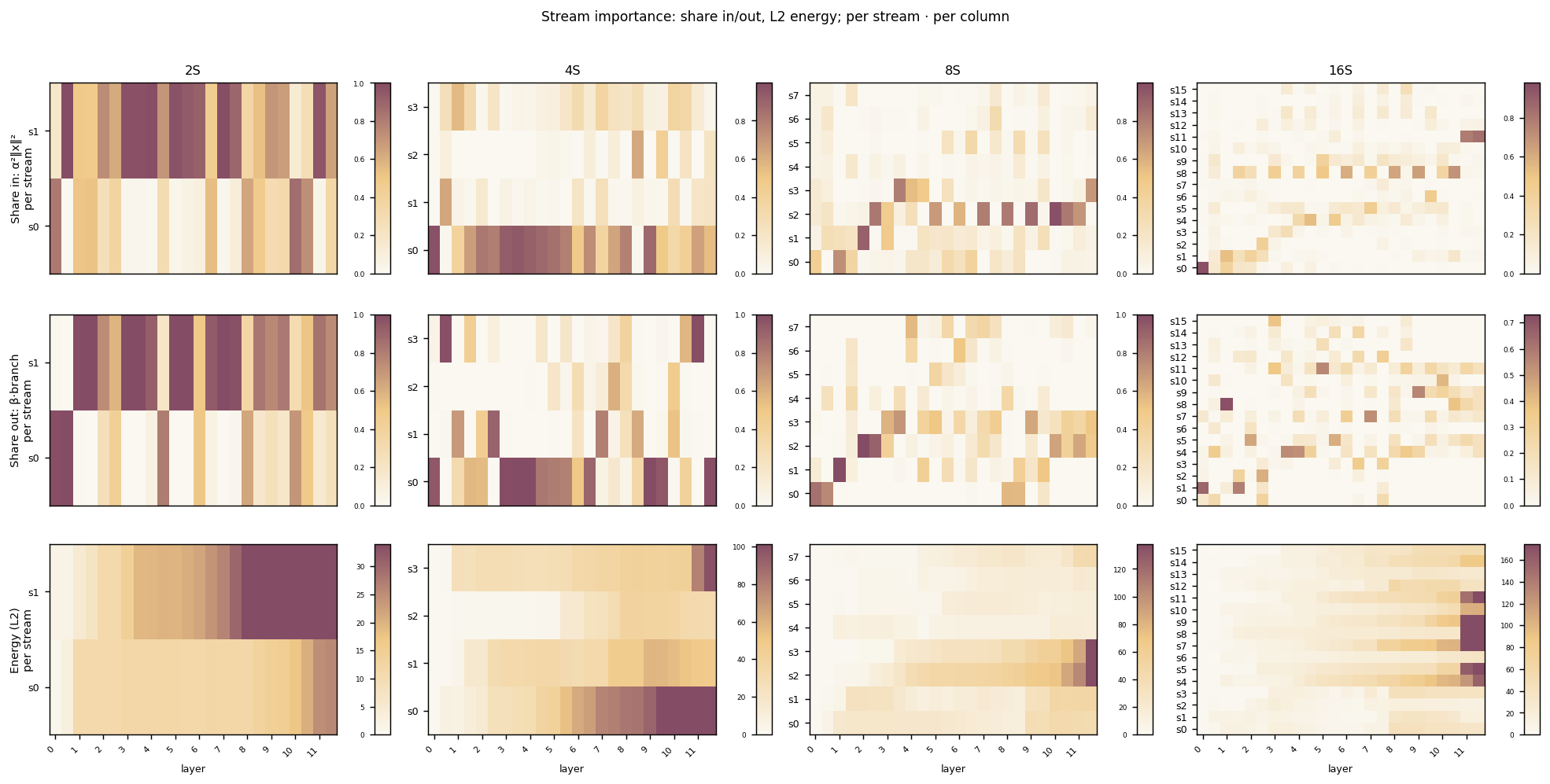}
  \caption{
  Stream-usage diagnostics for \textit{m}HC models with
  \(n\in\{2,4,8,16\}\) streams.
  Rows show normalized read share, normalized write share, and per-stream
  representation \(L_2\) norm across layers. Larger \(n\) spreads the signal
  across more streams, but the allocation remains nonuniform.
  }
  \label{fig:stream_count_importance}
\end{figure*}

Figure~\ref{fig:stream_count_importance} shows read share, write share, and
per-stream representation \(L_2\) norm across layers. The imbalance is visible
for all stream counts. It is strongest for \(n=2\), where one stream dominates
both routing and norm growth. For larger \(n\), especially \(n=16\), the signal
is distributed across more streams, but the allocation remains far from uniform.

\begin{figure*}[!htbp]
  \centering
  \includegraphics[width=0.7\textwidth]{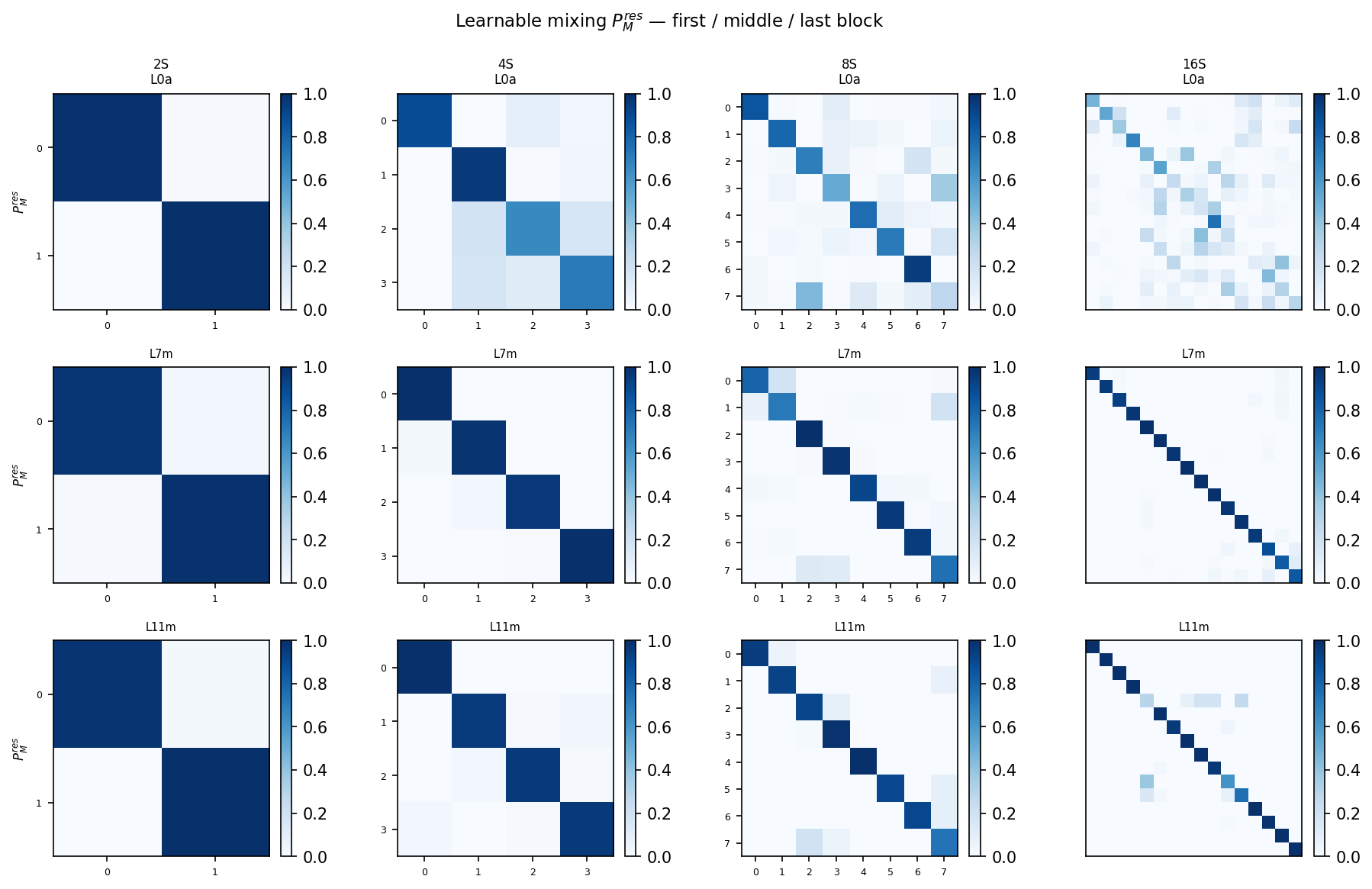}
  \caption{
  Representative residual-mixing matrices for \textit{m}HC models with
  \(n\in\{2,4,8,16\}\) streams.
  Residual mixing is strongest in the earliest block and becomes mostly
  near-diagonal in later layers across stream counts.
  }
  \label{fig:stream_count_mixing}
\end{figure*}

Figure~\ref{fig:stream_count_mixing} shows representative residual-mixing
matrices for the first, middle, and last blocks. Across stream counts, the
learned mixers become mostly near-diagonal after the earliest layers. Overall,
larger \(n\) softens but does not eliminate the imbalance: learned stream usage
remains nonuniform.

\section{Scaling across token budgets}
\label{sec:app_scaling}

Table~\ref{tab:lss_baseline_comparison} compares \textit{m}HC-lite and LSS
across token-to-parameter regimes corresponding to Chinchilla ratios
$0.5/1/2$. LSS improves perplexity consistently across OpenWebText,
WikiText-103, and C4, and the gains remain visible as the token budget
increases.

\begin{table}[htbp]
\centering
\caption{Perplexity ($\downarrow$) for \textit{m}HC-lite baseline and LSS
across token-to-parameter (T/P) rates.}
\label{tab:lss_baseline_comparison}
\small
\setlength{\tabcolsep}{4pt}
\begin{tabular}{c|c|cccc}
\toprule
T/P & Model & OWT val & WT val & WT test & C4 test \\
\midrule
10 & Base & 26.22 & 52.74 & 51.05 & 57.32 \\
10 & LSS  & \textbf{25.65} & \textbf{49.41} & \textbf{47.92} & \textbf{54.20} \\
\midrule
20 & Base & 24.16 & 48.59 & 46.32 & 52.09 \\
20 & LSS  & \textbf{23.63} & \textbf{45.66} & \textbf{43.57} & \textbf{50.39} \\
\midrule
40 & Base & 22.16 & 43.99 & 42.07 & 48.42 \\
40 & LSS  & \textbf{21.76} & \textbf{41.31} & \textbf{39.63} & \textbf{48.06} \\
\bottomrule
\end{tabular}
\end{table}

\section{Experimental details}
\label{sec:app_details}

Unless stated otherwise, ablations use the nanoGPT \textbf{M} model
(12 layers, $\sim$0.12B parameters) trained on OpenWebText for
$\approx 1.3$B tokens \citep{nanogpt, openwebtext}. We also evaluate the
nanoGPT \textbf{L} model (24 layers, $\sim$0.36B parameters). Following prior
HC work, we set the number of streams to $n=4$ \citep{mhc, mhc-lite}. We report
perplexity on OpenWebText validation, WikiText-103 validation/test, and C4 test
\citep{wikitext, c4}.

All experiments were run on NVIDIA H200 GPUs. As a reference point,
pre-training nanoGPT-M on 1.3B tokens takes approximately 6 GPU-hours, while
nanoGPT-L takes approximately 13 GPU-hours.
Seed-stability results are reported in Appendix~\ref{sec:stability}.

\subsection{Hyperparameters}

\begin{table}[htbp]
\centering
\small
\setlength{\tabcolsep}{6pt}
\begin{tabular}{lcc}
\toprule
\textbf{Hyperparameter} & \textbf{Medium} & \textbf{Large} \\
\midrule
\texttt{n\_layer}            & 12   & 24   \\
\texttt{n\_head}             & 12   & 16   \\
\texttt{n\_embd}             & 768  & 1024 \\
\texttt{dropout}             & 0.0  & 0.0  \\
\midrule
\texttt{learning\_rate}      & 6e-4 & 3e-4 \\
\texttt{min\_lr}             & 6e-5 & 3e-5 \\
\texttt{warmup\_iters}       & 2\%  & 2\%  \\
\midrule
\texttt{weight\_decay}       & 0.1  & 0.1  \\
\texttt{beta1}               & 0.9  & 0.9  \\
\texttt{beta2}               & 0.95 & 0.95 \\
\texttt{grad\_clip}          & 1.0  & 1.0  \\
\bottomrule
\end{tabular}
\vspace{2mm}

\caption{Training hyperparameters for nanoGPT Medium and Large runs.}
\label{tab:hparams_medium_large}
\end{table}

\newpage

\subsection{Notation}

\begin{table*}[htbp]
\centering
\small
\renewcommand{\arraystretch}{1.15}
\setlength{\tabcolsep}{6pt}
\begin{tabularx}{\textwidth}{@{} l l X @{}}
\toprule
\textbf{Symbol} & \textbf{Type / shape} & \textbf{Meaning} \\
\midrule

$\ell$ & index & Layer index. \\
$L$ & scalar & Number of layers. \\
$t$ & index & Token index. \\
$s$ & index & Stream index, typically $s\in\{1,\dots,n\}$. \\
$n$ & scalar & Number of parallel streams. \\
$d$ & scalar & Hidden dimension of each stream. \\

\midrule
$\mathbf{x}_\ell$ & $\in\mathbb{R}^{d}$ & Single-stream token representation at layer $\ell$; token index is often omitted. \\
$\mathbf{X}_\ell$ & $\in\mathbb{R}^{n\times d}$ & Multi-stream token state at layer $\ell$; rows correspond to streams. \\
$\mathbf{x}^{(s)}$ & $\in\mathbb{R}^{d}$ & Representation in stream $s$, i.e., a row/slice of $\mathbf{X}$. \\
$\mathbf{x}_t^{(\ell,s)}$ & $\in\mathbb{R}^{d}$ & Token $t$ representation at layer $\ell$ in stream $s$, used in cross-stream analysis. \\

\midrule
$\mathcal{F}_\ell(\cdot;\mathbf{W}_\ell)$ & function & Transformer block or residual block at layer $\ell$ with parameters $\mathbf{W}_\ell$. \\
$\mathbf{W}_\ell$ & parameters & Learnable parameters of block $\mathcal{F}_\ell$. \\

\midrule
$\mathbf{H}^{\mathrm{res}}_\ell$ & $\in\mathbb{R}^{n\times n}$ & Residual mixing operator: mixes the carried stream state in the residual branch. \\
$\mathbf{H}^{\mathrm{pre}}_\ell$ & $\in\mathbb{R}^{1\times n}$ & Read operator: forms the block input by reading from streams. \\
$\mathbf{H}^{\mathrm{post}}_\ell$ & $\in\mathbb{R}^{1\times n}$ & Write operator: distributes the block output back into streams. \\

\midrule
$\mathcal{B}_n$ & set & Birkhoff polytope: set of doubly stochastic $n\times n$ matrices. \\
$\mathbf{1}$ & vector & All-ones vector used in doubly stochastic constraints. \\
$\mathbf{P}$ & $\in\mathbb{R}^{n\times n}$ & Permutation matrix acting on the stream axis. \\
$\mathbf{E}(\mathbf{x})$ & $\in\mathbb{R}^{n\times d}$ & Standard HC expansion by replication, $\mathbf{E}(\mathbf{x})=[\mathbf{x};\ldots;\mathbf{x}]$. \\

\midrule
$\mathbf{h}_s$ & $\in\mathbb{R}^{d}$ & Per-stream diagonal scaling vector used in Learned Stream Scaling. \\
$\odot$ & operator & Elementwise product. \\

\midrule
$\mathbf{a}_t^{(\ell)}$ & $\in\mathbb{R}^{m}$ & Sparse activation/code for token $t$ in the crosscoder at layer $\ell$. \\
$m$ & scalar & Number of sparse features in the crosscoder dictionary. \\
$\mathbf{D}^{(\ell,s)}$ & $\in\mathbb{R}^{d\times m}$ & Decoder matrix for layer $\ell$, stream $s$. \\
$\hat{\mathbf{x}}_t^{(\ell,s)}$ & $\in\mathbb{R}^{d}$ & Crosscoder reconstruction, $\hat{\mathbf{x}}_t^{(\ell,s)}=\mathbf{D}^{(\ell,s)}\mathbf{a}_t^{(\ell)}$. \\
$i$ & index & Sparse feature index. \\
$w_i^{(\ell,s)}$ & scalar & Decoder-column norm, $w_i^{(\ell,s)}=\lVert \mathbf{D}^{(\ell,s)}_{:,i}\rVert_2$. \\
$s_i^{(\ell)}$ & index & Stream assigned to feature $i$ at layer $\ell$, $s_i^{(\ell)}=\arg\max_s w_i^{(\ell,s)}$. \\

\bottomrule
\end{tabularx}
\caption{Notation used throughout the paper.}
\label{tab:notation}
\end{table*}

\newpage
\section{Training Stability Across Random Seeds}
\label{sec:stability}

\begin{figure}[htbp]
  \centering
  \includegraphics[width=\columnwidth]{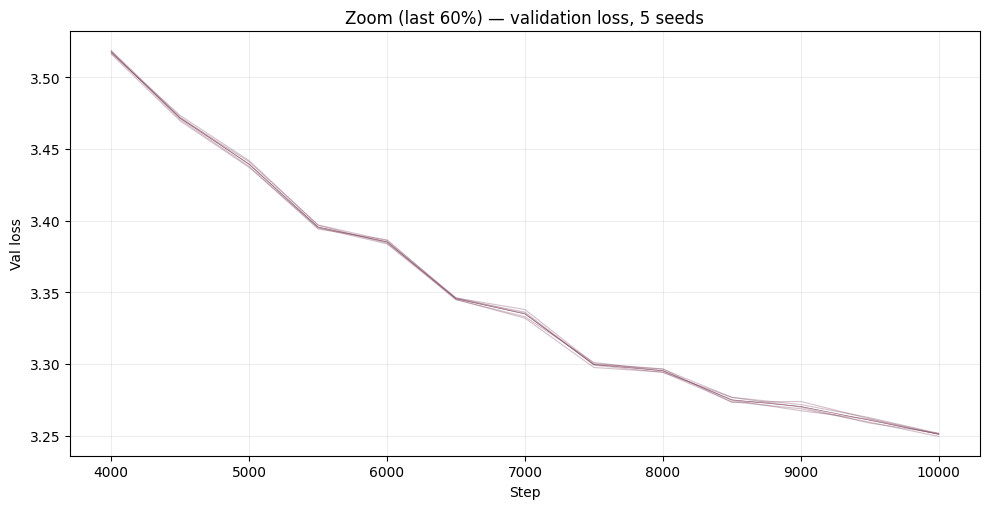}
  \vspace{-2mm}
  \caption{Validation loss across 5 random seeds (zoomed to the last 60\% of training), showing tightly clustered trajectories and low seed sensitivity.}
  \label{fig:stability}
  \vspace{-4mm}
\end{figure}

We evaluated training stability of medium \textit{m}HC-lite model by repeating the same setup with 5 different random seeds and comparing the resulting training dynamics.
Figure~\ref{fig:stability} shows the validation-loss trajectories over the later part of training (zoomed to emphasize small differences).

Across seeds, the curves are tightly clustered throughout training: the end-of-training validation loss varies within a narrow range (final min--max range $\approx 3.2\times 10^{-3}$; final std $\approx 8.8\times 10^{-4}$), and typical pairwise relative deviations stay below $\sim 0.15\%$.

Overall, we observe low sensitivity to random initialization and no divergent runs, suggesting that the qualitative conclusions of our experiments are robust and not driven by seed-specific effects.

\section{Cross-stream mixing across token budgets}
\label{sec:app_mixing}

\begin{figure}[p]
\centering
\includegraphics[height=0.49\textheight,keepaspectratio]{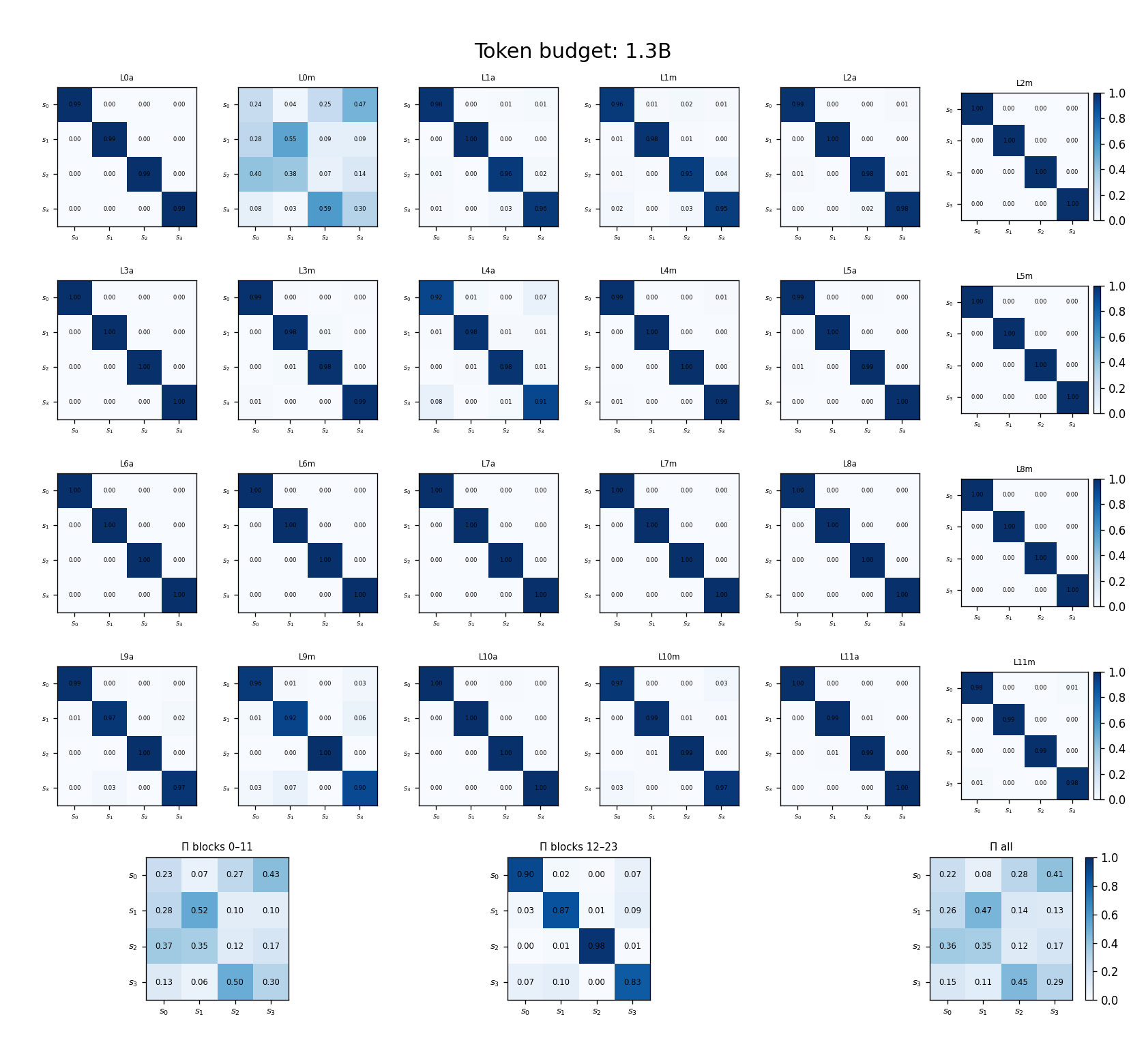}
\vspace{6pt}

\includegraphics[height=0.49\textheight,keepaspectratio]{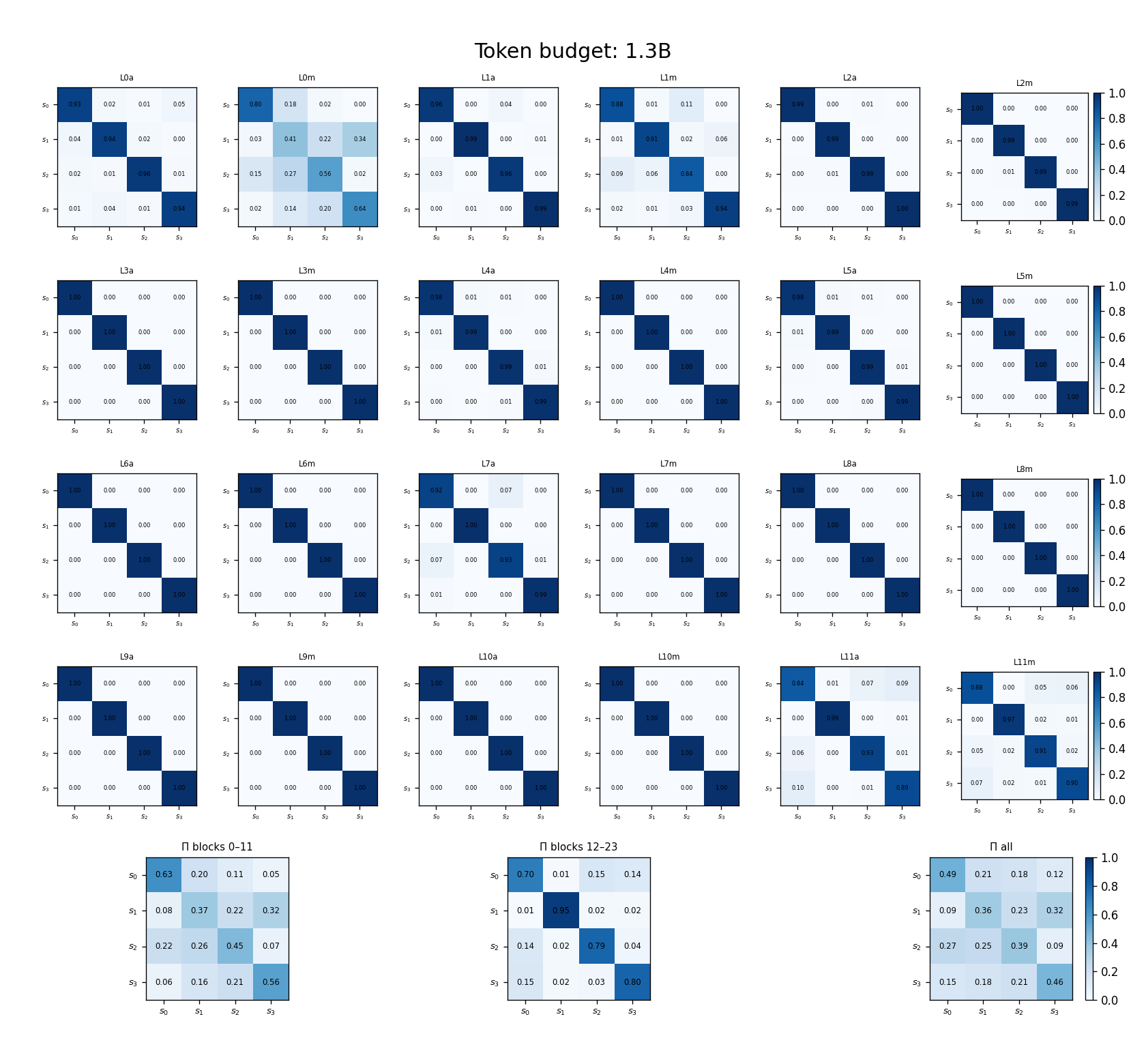}

\caption{Token budget 1.3B: \textit{m}HC-lite (top), LSS (bottom).}
\label{fig:tok_13}
\end{figure}

\begin{figure}[p]
\centering
\includegraphics[height=0.49\textheight,keepaspectratio]{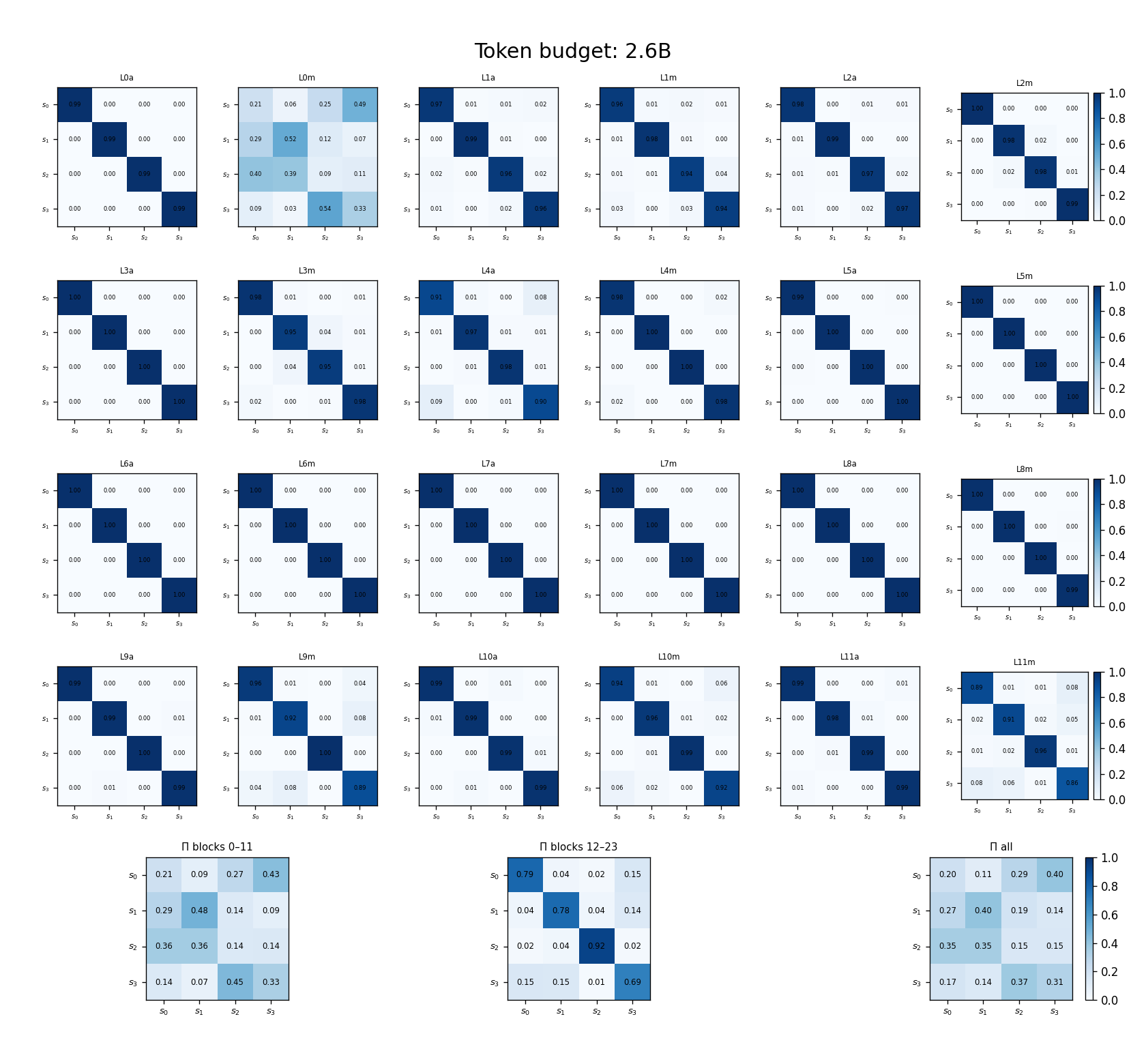}

\vspace{6pt}

\includegraphics[height=0.49\textheight,keepaspectratio]{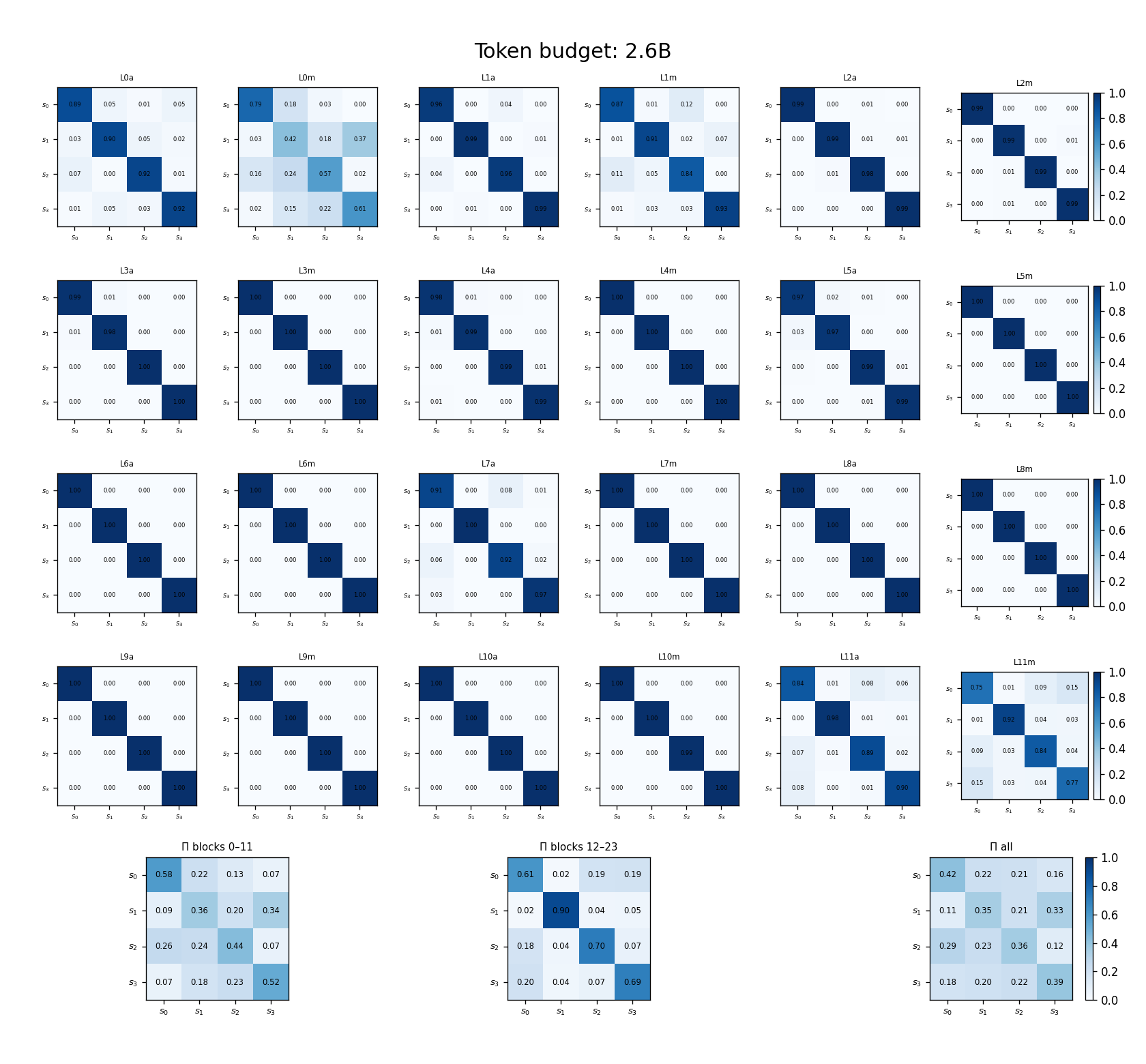}
\caption{Token budget 2.6B: \textit{m}HC-lite (top), LSS (bottom).}
\label{fig:tok_26}
\end{figure}

\begin{figure}[p]
\centering
\includegraphics[height=0.49\textheight,keepaspectratio]{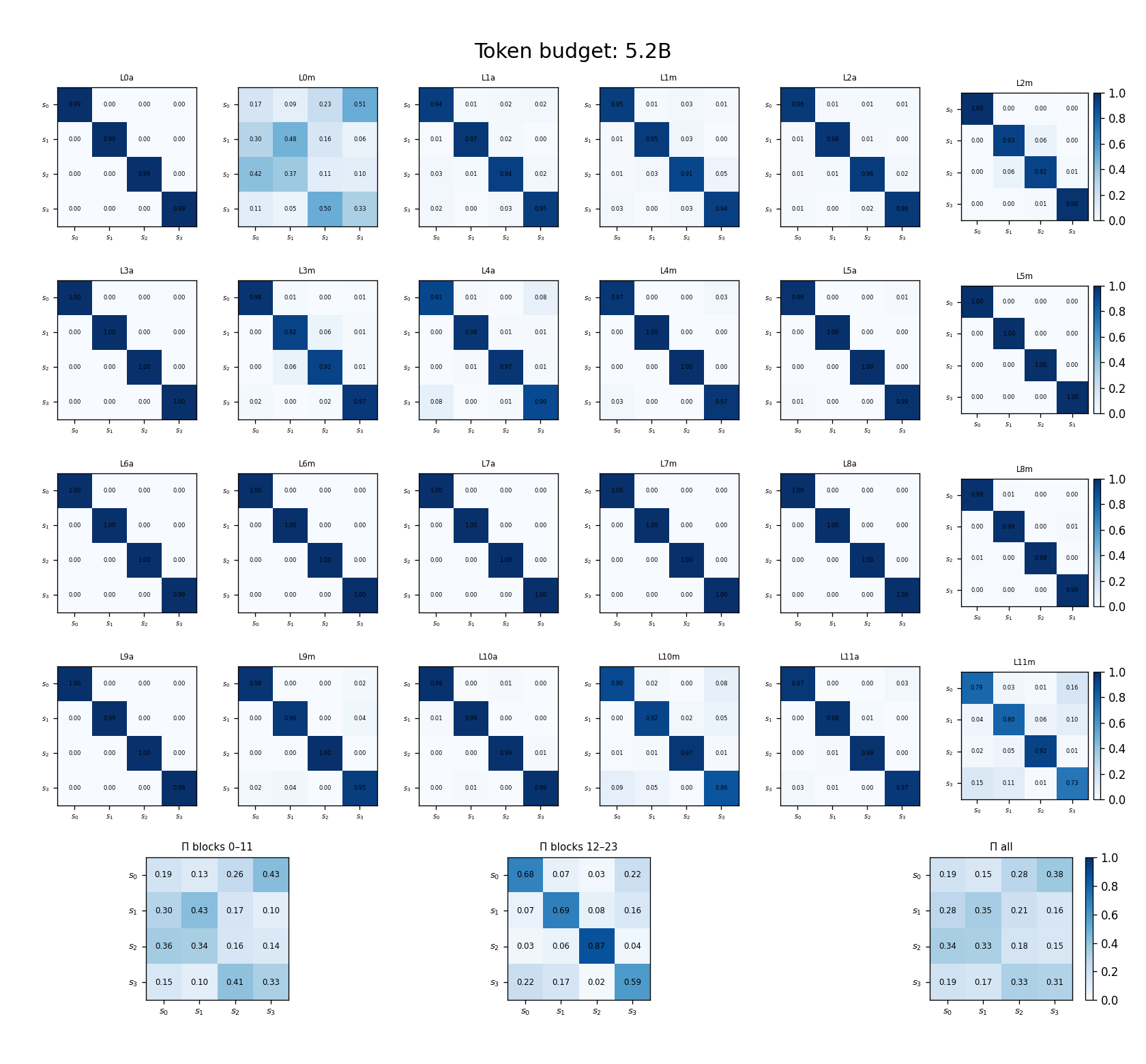}

\vspace{6pt}

\includegraphics[height=0.49\textheight,keepaspectratio]{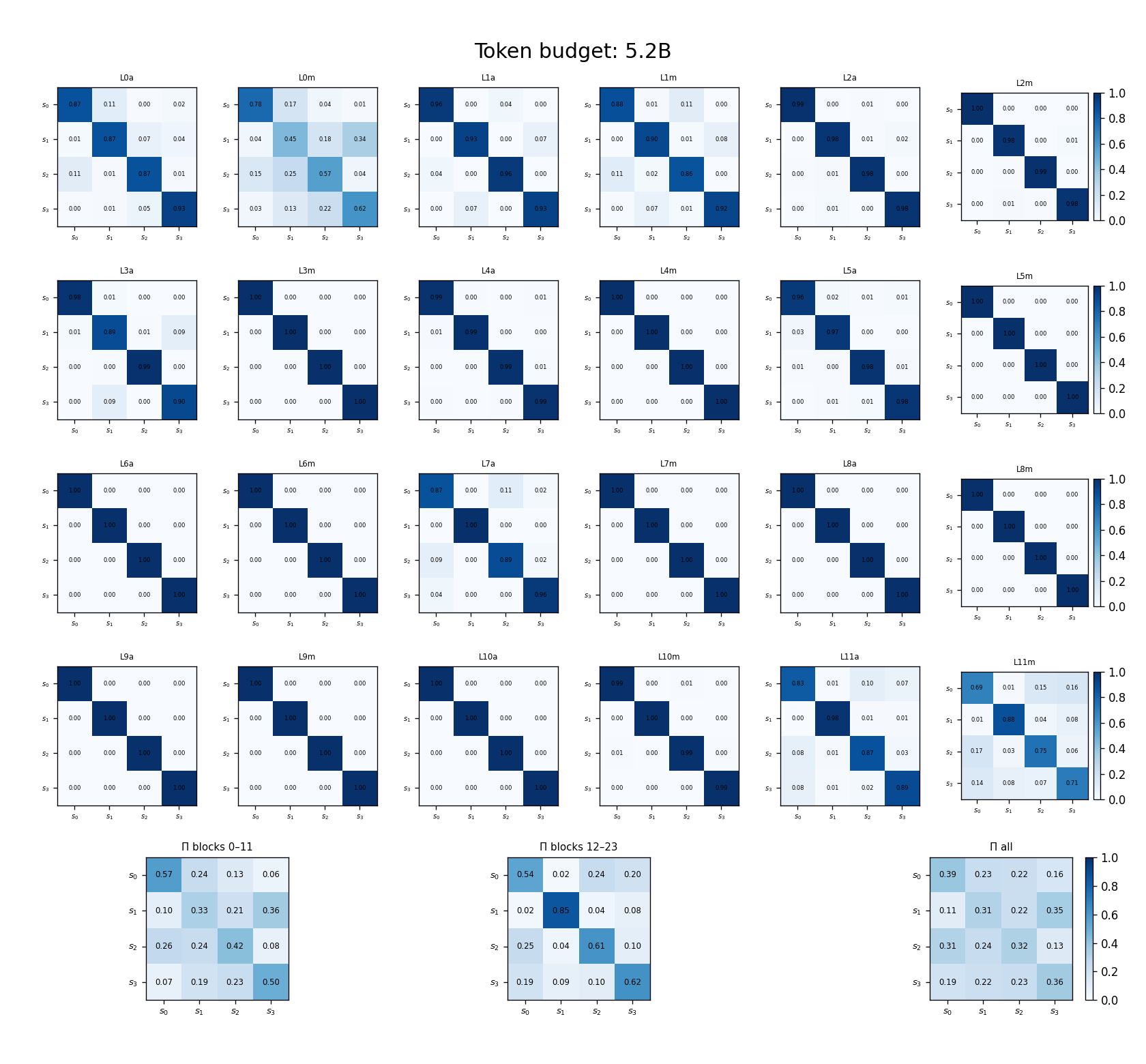}
\caption{Token budget 5.2B: \textit{m}HC-lite (top), LSS (bottom).}
\label{fig:tok_52}
\end{figure}

In this appendix we visualize the learned residual-mixing operators across depth for two variants:
the standard \textit{m}HC-lite baseline and the LSS variant.
Each page corresponds to a different token budget (1.3B / 2.6B / 5.2B), and shows the per-layer
mixing heatmaps for the two models. See Figures \ref{fig:tok_13} - \ref{fig:tok_52} for details.

Overall, strong cross-stream mixing emerges very slowly: for most layers the learned mixing remains
close to an identity-like pattern, indicating that sustained exchange between streams is limited and
builds up only gradually with training budget. At the same time, compared to the baseline,
our scheme starts to exhibit noticeable off-diagonal mass earlier, i.e., the onset of stream
mixing appears sooner. Nevertheless, even with larger budgets, the effect remains modest and the
dominant regime is still characterized by weak/slowly developing mixing.

\clearpage

\clearpage

\clearpage

\end{document}